# Towards Aggregating Weighted Feature Attributions

**Umang Bhatt**
Carnegie Mellon University
umang@cmu.edu

**Pradeep Ravikumar**
Carnegie Mellon University
pradeepr@cs.cmu.edu

**José M. F. Moura**
Carnegie Mellon University
moura@cmu.edu

## Abstract

Current approaches for explaining machine learning models fall into two distinct classes: antecedent event influence and value attribution. The former leverages training instances to describe how much influence a training point exerts on a test point, while the latter attempts to attribute value to the features most pertinent to a given prediction. In this work, we discuss an algorithm, AVA: Aggregate Valuation of Antecedents, that fuses these two explanation classes to form a new approach to feature attribution that not only retrieves local explanations but also captures global patterns learned by a model. Our experimentation convincingly favors weighting and aggregating feature attributions via AVA.

## Introduction

Over a quarter century ago, Herbert Simon, a pioneer of artificial intelligence, articulated the intricacies of explaining human behavior (Simon 1992). He proposed that, since humans behave in a rational manner, their actions can be explained through a "fugue [composition], with two intertwined themes" – one theme comes from examining a human's historical behavior and the other comes from the laws and factors at play in a human's environment. These themes naturally give rise to human behavior explanations. The first type of explanation is what we can call **antecedent event influence**: using past events to describe the present. By referencing back to antecedent events, humans can justify the motives and intents behind their behaviors and choices in a given event. The second type of explanation is **value attribution**. An explanation by antecedent events also would require humans to have a deep understanding of the factors and laws driving each action. Humans consider multiple factors and complex associations between those factors, when explaining their own behavior.

Machine learning at some level of abstraction could be viewed as explicitly grappling with these two themes: looking at past events as captured by training data, deriving an understanding of how various environmental variables are associated with each other, and finally associating the two with a target response. Accordingly, these two themes could be used to *explain* machine learning models in two different ways. Antecedent event influence reduces to explaining a model's output by looking at the training examples most influential to the instance being classified (Koh and Liang 2017). In a machine learning context, value attribution would ideally provide a rich associational or even causal structure among all the variables in the system. Unfortunately, this is too difficult to obtain, especially given access to just a black-box model. Recent approaches for value attribution loosely provide associations between input variables and a target response, i.e., a ranking of influential input variables for a target prediction (Lundberg and Lee 2017). Unfortunately, this is a relatively impoverished notion of value attribution that does not provide us with a rich understanding of global model behavior.

In this work, we study model interpretability by building a post-hoc explanation algorithm that combines these two *themes*. We restrict ourselves to post-hoc explanations from already learned black-box models (Lipton 2016). Our method, called AVA, combines existing approaches for antecedent event influence and (impoverished) value attribution to provide an explanation in the form of a consensus value attribution. Specifically, we aggregate the value (or feature) attributions of a test point's $k$ most influential training points to capture a local explanation for the test point and find global trends learned by the model.

## Related Work

We provide a brief review of recent approaches along each of the two themes outlined above: antecedent event influence and value attribution.

### Antecedent event influence

Antecedent event influence techniques are limited in machine learning, mostly due to memory and computational limitations of retrieving appropriate training data points. The simplest class of such approaches are based on obtaining those data points in the training data that are most similar to the test data point. The caveat is that this relies on a suitable notion of similarity, which is non-trivial to specify for complex domains. A recently proposed class of techniques (Koh and Liang 2017) use influence functions, a classic tool from robust statistics, to tractably obtain such influential training points. Suppose the training loss function is $\mathcal{L}$, the learned predictor given the training data is $\widehat{f}$, and we are interested in the counterfactual of how upweighting a training data point $x$ infinitesimally will affect the loss at a test point $x_{\text{test}}$. Denoting the predictor obtained by upweighting training data point $x$ by some constant $\epsilon$ as $\widehat{f}_{\epsilon,x}$, the influence function based approach quantifies the influence of a training point $x$, on the loss incurred by the model $\widehat{f}$ at a test point $x_{\text{test}}$, as follows.

$$\mathcal{I}_{\text{up,loss}}(x, x_{\text{test}}) = \frac{d}{d\epsilon}\mathcal{L}(\widehat{f}_{\epsilon,x}, x_{\text{test}})\big|_{\epsilon=0}$$

Koh and Liang provide relatively tractable approximation scheme to approximate the above (Koh and Liang 2017).

**Value attribution**

On the other hand, value attribution has attracted considerably greater attention, due in part to its greater tractability with respect to both computation and storage. Parametric models such as linear regression models are precisely parameterized by such input feature weights. Such simple parametric models could be used to derive value attribution by approximating the given model either globally or locally.

A recent line of work that modifies such simple gradients to provide improved explanations. Among these methods lie Integrated Gradients (IG) (Sundararajan, Taly, and Yan 2017), which compute the average gradient while the input varies along a linear path to $x$ starting at a generally zero baseline $\bar{x}$, so that the $i$-th coordinate of the explanation at a test point $x$ is given as follows.

$$IG_i(x) = (x_i - \bar{x}_i)\int_{\alpha=0}^{1} \frac{\partial f(\bar{x} + \alpha(x - \bar{x}))}{\partial x_i}d\alpha$$

Another class of measures draw from game-theory and revenue division based on various approaches to address the following question: if we can compute the marginal contribution of a given feature, given some subset of the set of features (e.g., by comparing the predictor trained on just the subset vs the predictor trained on the subset with the feature added), then what is the overall marginal influence $\phi_i$ of a feature $i$? The most popular approach to do so is Shapley values (Shapley 1953), which uses the entire powerset of the set of all features $F$, and which when used in an explanation context has been termed SHapley Additive exPlanations (SHAP) (Lundberg and Lee 2017):

$$\phi_i = \sum_{S \subseteq F\setminus\{i\}} R\left(f_T(x_T) - f_S(x_S)\right)$$

where $R = \left(\frac{|S|!(|F|-|S|-1)!}{|F|!}\right)$, $T = S \cup \{i\}$, and $f_T$ is the model predictor only trained with features in subset $T$.

There is thus a burgeoning line of work along both antecedent event influence and value attribution, particularly the latter. As noted in the introduction, the current set of approaches for value attribution are relatively impoverished, focusing largely on associations of input features with a target response. To address this, we propose Aggregate Valuation of Antecedents, AVA, which is a new class of explanations that aims to fuse antecedent event influence and value attribution, while exposing global patterns learned by the model.

## Aggregate Valuation of Antecedents

We set up our approach with an example. Imagine a doctor attempting to decide whether each of her ten patients has the flu or not based off of three factors (heart rate, blood pressure, body temperature). The collection of patients can be represented as a set $D = \{X_1, X_2 \ldots X_{10}\}$ where each $X_i \in \mathbb{R}^3$. The doctor can assign one of $c$ different ICD-9 codes. The doctor can look at data from their previous patients most similar to the one they are currently examining: in machine learning, we can abstract this to the training set of a model, where the patients that come to mind are the most influential in diagnosing the current patient, our unobserved test point. Moreover, the doctor can use information about which of the three factors were relevant in previous diagnoses.

### Approach

Our proposed method, AVA, combines the idea of approximating a black-box model of interest to develop a value attribution for a test point via (Lundberg and Lee 2017; Sundararajan, Taly, and Yan 2017) with a local neighborhood influence measure proposed in (Koh and Liang 2017).

First let's introduce some notation. Let $x \in \mathbb{R}^d$ be a datapoint's feature vector where the $x_i \in \mathbb{R}$ is a specific feature of this datapoint. Let $\mathcal{D} = \{x^{(j)}\}_{j=1}^N$ represent the training datapoints, where $\mathcal{D} \in \mathbb{R}^{d \times N}$ is the entire training set in matrix form with $\mathcal{D}_{i,j} = x_i^{(j)}$. Let $\widehat{f}$ be the learned predictor we wish to explain. We define the influence weight, $\rho_j$ of training point, $x^{(j)}$, on a test point, $x_{\text{test}}$ as:

$$\rho_j = \mathcal{I}_{\text{up,loss}}(x^{(j)}, x_{\text{test}}) = \frac{d}{d\epsilon}\mathcal{L}(\widehat{f}_{\epsilon,x^{(j)}}, x_{\text{test}})\big|_{\epsilon=0}$$

We define $\{\rho_j\}_{x^{(j)} \in \mathcal{D}} \in \mathbb{R}^N$ as the set of influence weights for all training points where $\rho_i \in \mathbb{R}_{\geq 0}$. These influences weights induce a probability distribution over the feature space centered at $x_{\text{test}}$, where every training point exerts a non-negative weight on the test point. Next, we select the local neighborhood, $\mathcal{N}_k$, of the $k$ most influential training points on $x_{\text{test}}$:

$$\mathcal{N}_k(x_{\text{test}}, \mathcal{D}) = \arg\max_{\mathcal{N} \subset \mathcal{D}, |\mathcal{N}|=k} \sum_{x^{(j)} \in \mathcal{N}} \rho_j$$

We define $\{\rho_j\}_{x^{(j)} \in \mathcal{N}_k} \in \mathbb{R}^k$, as the set of influence weights for the top $k$ influential points. Using an attribution technique $g$ (SHAP, IG, etc.), we obtain a value attribution for each of the $k$ points. Let $g^j$ be the attribution vector of the $j^{th}$ influential training point and $g_i^j$ be the attribution of feature $i$ of that point. Finally, once we have the set of value attributions $\{g^j\}_{x^{(j)} \in \mathcal{N}_k} \in \mathcal{G}^*$, where each $g^j \in \mathcal{G}$, we can apply an aggregation scheme: $\mathcal{A} : \mathcal{G}^* \mapsto \mathcal{G}$ to obtain a consensus value attribution.

### Aggregation

In order to build a sound attribution technique, AVA relies on a clever aggregation mechanism, $\mathcal{A}$. For each attribution technique (SHAP and IG), we find a suitable aggregation mechanism, $\mathcal{A}_{\text{SHAP}}$ and $\mathcal{A}_{\text{IG}}$ respectively.

**SHAP** If we let $g(x)$ be SHAP attribution, we find that attribution of the $i^{th}$ feature of point $x$ is given by the Shapley value, which is the sum of the contributions to $\widehat{f}$ for the $i^{th}$ feature in all possible subsets $S$ of the features $F$ given by:

$$\phi_i(x) = \sum_{S \subseteq F \setminus \{i\}} R\left(\widehat{f}_{S \cup \{i\}}(x_{S \cup \{i\}}) - \widehat{f}_S(x_S)\right)$$

We let $R = \left(\frac{|S|!(|F|-|S|-1)!}{|F|!}\right)$. Alternatively, following suit of (Yokote, Funaki, and Kamijo 2015), we can write Shapley value feature importance as:

$$\phi_i(x) = \sum_{S \subseteq F \setminus \{i\}} \frac{1}{|S|} D(\widehat{f}, S, x)$$

where $D$ is defined as the Harsanyi dividend, a measure of how much a given subset of all features contributes to the final feature attribution:

$$D(\widehat{f}, S, x) = \sum_{T \subseteq S} (-1)^{|S \setminus T|} \widehat{f}(x_{S \cup \{i\}})$$

Shapley values satisfy additivity (Shapley 1953) across games; as such, we know the following holds: $\phi_i(x_1 + x_2) = \phi_i(x_1) + \phi_i(x_2)$ for independent games $x_1$ and $x_2$. For us, this means that we can combine Shapley value contributions of features across different datapoints. Moreover, Shapley values satisfy scaling as well: $\phi_i(cx_3) = c\phi_i(x_3)$, which allows us to weigh each point by its corresponding influence. (Kalai and Samet 1987) proposed the weighted Shapley value which would weigh every dividend by a player's weight. In our case, we weigh each feature's contribution from every influential point $(x^{(j)})$ by its influence weight $(\rho_j)$ as follows.

$$g_i(x^{(j)}) = \phi_i(x^{(j)}) = \sum_{S \subseteq F \setminus \{i\}} \frac{\rho_j}{\rho|S|} D(\widehat{f}, S, x^{(j)})$$

Let $\rho = \sum_{i \in S} \rho_i$. Since $g_i(x^{(j)})$ represents the weighted attribution of the $i^{th}$ feature in the $j^{th}$ datapoint and since Shapley values allow for scaling and additivity, we sum the weighted feature attributions across all influential datapoints.

$$\mathcal{A}_{\text{SHAP}}(\{g(x)\}_{x \in \mathcal{N}_k}) = \sum_{x^{(j)} \in \mathcal{N}_k} \sum_{S \subseteq F \setminus \{i\}} \frac{\rho_j D(\widehat{f}, S, x^j)}{\rho|S|}$$

We simplify to get the following where $g^j$ is an attribution vector of weighted Shapley values $g_i$ for all $d$ features of the $j^{th}$ influential datapoint.

$$\mathcal{A}_{\text{SHAP}}(\{g(x)\}_{x \in \mathcal{N}_k}) = \sum_{x^{(j)} \in \mathcal{N}_k} \frac{\rho_j}{\rho} g^j$$

**Integrated Gradients** If we let $g(x)$ be IG attribution, we find that attribution of the $i^{th}$ feature of point $x$ is given by the gradient of $\widehat{f}(x)$ along the $i^{th}$ dimension of $x$ with respect to a baseline $\bar{x}$.

$$g_i(x) = (x_i - \bar{x}_i) \int_{\alpha=0}^{1} \frac{\partial \widehat{f}(\bar{x} + \alpha(x - \bar{x}))}{\partial x_i} d\alpha$$

However, IG lets the baseline, $\bar{x}$, be an input vector with no output signal or just the zero vector: whatever satisfies $\widehat{f}(\bar{x}) \approx 0$. Though this provides satisfactory attributions in isolation, it neglects any global trends learned by the model. AVA alters the baseline to be the influence-weighted average of the local neighborhood, $\mathcal{N}_k$. Recall we assume black-box access to our learned predictor $\widehat{f}$ and do not modify its internals. As such, we desire AVA with IG attribution to consider straightline paths in $\mathbb{R}^d$ from a baseline of the influence-weighted average of the local neighborhood to the point $x_{\text{test}}$, computes the gradients at all the influential points along the path, and finds an attribution by accumulating these gradients. Concretely, AVA weighs and aggregates the path integrals of the gradients along the straightline path from an arbitrary baseline $x_0$ to the $k$ influential points.

Similar to weighted Shapley attribution, we take $\mathcal{A}_{\text{IG}}$ to be the following, where we let $\rho = \sum_{x^{(i)} \in \mathcal{N}_k} \rho_i$.

$$\mathcal{A}_{\text{IG}}(\{g(x)\}_{x \in \mathcal{N}_k}) = \sum_{x^{(j)} \in \mathcal{N}_k} \frac{\rho_j}{\rho} g^j$$

## Experiments

In this section, we present experiments to evaluate the consensus attribution given by AVA on tabular datasets, showing not only the fidelity of its local explanations but also the appearance of global patterns. For our experiments, we leverage several UCI datasets (Dheeru and Karra Taniskidou 2017). We divide each dataset into train and test with a 33% split.

### Local Explanations

To explain an individual prediction via value attribution, we compare AVA with the attribution given by the feature attribution technique itself (SHAP or IG).

**Gold Set Recall** We can quantify the faithfulness of a value attribution through its recall on a *gold set* of $m$ important features obtained from an interpretable model (Ribeiro, Singh, and Guestrin 2016). To obtain a *gold set*, we use a decision tree classifier that we prune to a maximum of $m$ features, where $m$ is picked by cross validation for each dataset to maximize accuracy of the known-interpretable classifier; in practice, $m$ ought to be selected by the user to determine the complexity of the explanation provided. As a sanity check, we also compare against a random procedure that randomly picks $m$ features as an explanation.

| Dataset | Random | SHAP | IG | $\mathcal{A}_{\text{SHAP}}$ | $\mathcal{A}_{\text{IG}}$ |
|---|---|---|---|---|---|
| Adult | 50 | 78 | 63 | **81** | 74 |
| Communities | 45 | 58 | 68 | 61 | **71** |
| Diabetes | 25 | 40 | 47 | **50** | **50** |
| Iris | 50 | 89 | 64 | **91** | 67 |
| Titanic | 37 | 55 | 63 | 60 | **88** |

Table 1: Gold set recall on important features from an interpretable classifier to explain a three layer MLP with sigmoid activation and trained with ADAM

| DATASET | RANDOM | SHAP | IG | $\mathcal{A}_{\text{SHAP}}$ | $\mathcal{A}_{\text{IG}}$ |
|---|---|---|---|---|---|
| ADULT | 50 | 76 | 56 | **82** | 65 |
| COMMUNITIES | 51 | 59 | 65 | **71** | 71 |
| DIABETES | 25 | 22 | 39 | 49 | **50** |
| IRIS | 50 | 64 | 18 | **72** | 24 |
| TITANIC | 37 | 78 | 70 | 75 | **95** |

Table 2: Gold set recall on important features from an interpretable classifier to explain a three layer MLP with ReLu activation and trained with ADAM

| DATASET | RANDOM | SHAP | IG | $\mathcal{A}_{\text{SHAP}}$ | $\mathcal{A}_{\text{IG}}$ |
|---|---|---|---|---|---|
| ADULT | 50 | 78 | 65 | **80** | 73 |
| COMMUNITIES | 47 | 41 | 60 | 34 | **67** |
| DIABETES | 24 | 27 | **50** | 29 | 50 |
| IRIS | 50 | 80 | 62 | **86** | 70 |
| TITANIC | 37 | 72 | 66 | 75 | **96** |

Table 3: Gold set recall on important features from an interpretable classifier to explain an SVM with an RBF kernel

| DATASET | RANDOM | SHAP | IG | $\mathcal{A}_{\text{SHAP}}$ | $\mathcal{A}_{\text{IG}}$ |
|---|---|---|---|---|---|
| ADULT | 50 | 77 | 64 | **81** | 70 |
| COMMUNITIES | 45 | 33 | 49 | 50 | **73** |
| DIABETES | 25 | 30 | 33 | 31 | **50** |
| IRIS | 50 | **95** | 18 | 85 | 26 |
| TITANIC | 37 | 70 | 71 | **99** | 76 |

Table 4: Gold set recall on important features from an interpretable classifier to explain a kNN classifier

We use SHAP, IG, and AVA (with both attribution techniques) to explain an MLP and report recall of a decision tree's *gold set* averaged over all the test instances for multiple datasets in Table 1 and Table 2. As a sanity check, we also report the recall when explaining a SVM with an RBF kernel and a kNN classifier in Table 3 and Table 4 respectively. For these experiments, we fix $k$ to be ten, use the mean values of the training input as the region of perturbation for SHAP, and use the aforementioned greedy technique to determine $m$. Note random attribution will recall $m/d$, where $d$ is the total number of features. On all datasets, AVA consistently outperforms or matches SHAP or IG alone for a test point's feature attribution.

$k$-**Sensitivity** The local fidelity of AVA extends to how sensitive AVA is to $k$, the number influences aggregated. We find that AVA remains relatively constant and stable as we increase $k$. In Figure 1, when trying explaining a three layered MLP with sigmoid activations (trained with ADAM) for two different datasets, we see that AVA outperforms SHAP and IG and find that after eight influences, we stabilize recall. The graph portrays what percentage gold set recall (y-axis) as a function of the number of influences used $k$ (x-axis).

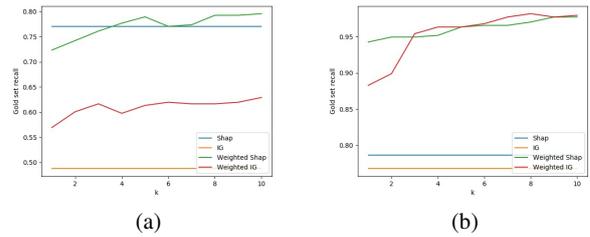

Figure 1: $k$-Sensitivity of AVA: (a) Adult; (b) Titanic

**Global Patterns**

Not only does AVA more accurately recover local explanations via its consensus attribution, but AVA also identifies global trends in the data. AVA extracts how much weight each attribution scheme gives to the top $m$ features; as such, we calculate the **mean feature importance** of the top $m$ features. If all the information about a prediction is contained in the top $m$ features, then we expect the mean feature importance to be the upper bound, $1/m$; if feature importance is uniformly distributed over all features, we expect the mean information to be the lower bound, $1/d$. Note a uniform distribution of feature importance would be uninformative in deciding which feature is important to $\widehat{f}$, so we desire feature importance to be asymmetrically distributed. We report the mean feature importance of different attribution techniques for the Adult and Titanic datasets in Figure 2: the higher, the better. The line represents the upper bound of mean feature importance. We conclude AVA with either attribution provides a more effective attribution than current techniques.

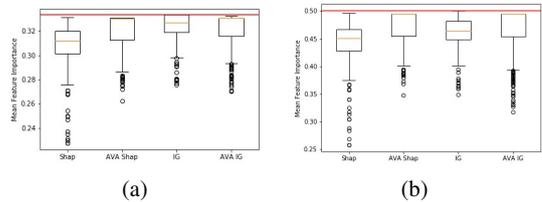

Figure 2: Mean feature importance in a three layer MLP with ReLu activation for: (a) Adult; (b) Titanic

## Conclusion

In this paper, we discussed AVA, Aggregate Valuation of Antecedents, as a value attribution technique that not only gives local explanations but also detect global patterns. By calculating the top $k$ influences for a given test point, we aggregate those influences' feature attributions to find a consensus attribution. We have shown that AVA's consensus attribution outperforms current attribution benchmarks on tabular datasets. In future work, we hope to realize a medical use case of AVA, develop a more robust aggregation step that builds on counterfactual intuition, and adapt AVA for unstructured domains (i.e., images and natural language).